\documentclass{article}
\usepackage{spconf,amsmath,graphicx,amsthm,amssymb}

\title{Self-critical Sequence Training for Automatic Speech Recognition}
\name{Chen Chen, Yuchen Hu, Nana Hou, Xiaofeng Qi, Heqing Zou, Eng Siong Chng
\sthanks{This research is supported by National Research Foundation Singapore under its AI Singapore Programme (Award Number: AISG-100E-2018-006).}}
\address{School of Computer Science and Engineering, Nanyang Technological University, Singapore \\   chen1436@e.ntu.edu.sg}
\begin{document}
%
\maketitle
\begin{abstract}
Although automatic speech recognition (ASR) task has gained remarkable success by sequence-to-sequence models, there are two main mismatches between its training and testing that might lead to performance degradation: 1) The typically used cross-entropy criterion aims to maximize log-likelihood of the training data, while the performance is evaluated by word error rate (WER), not log-likelihood; 2) The teacher-forcing method leads to the dependence on ground truth during training, which means that model has never been exposed to its own prediction before testing. In this paper, we propose an optimization method called self-critical sequence training (SCST) to make the training procedure much closer to the testing phase. As a reinforcement learning (RL) based method, SCST utilizes a customized reward function to associate the training criterion and WER. Furthermore, it removes the reliance on teacher-forcing and harmonizes the model with respect to its inference procedure. We conducted experiments on both clean and noisy speech datasets, and the results show that the proposed SCST respectively achieves 8.7\% and 7.8\% relative improvements over the baseline in terms of WER.  
\end{abstract}
\begin{keywords}
Automatic Speech Recognition, Reinforcement leaning 
\end{keywords}
\section{Introduction}
\label{sec:intro}
In recent years, sequence-to-sequence automatic speech recognition (ASR) models, such as CTC \cite{kim2017joint}, LAS \cite{chan2016listen}, RNN-T \cite{graves2013speech}, and Transformer \cite{mohamed2019transformers}, have achieved impressive performance and dominate the leader-board. However, these models still suffer from mismatches between their training and testing stages. \par
The first mismatch is between the training objective and the testing metric. Most sequence-to-sequence model is typically optimized by the cross-entropy (CE) criteria, which corresponds to maximizing the log-likelihood for each frame \cite{prabhavalkar2018minimum}. However, during testing, the metric to evaluate a trained model is the task-specific criteria, such as the word error rate (WER), not log-likelihood. This discrepancy might lead to sub-optimal performance in terms of WER. The second mismatch is caused by the widely used teacher-forcing \cite{williams1989learning} during training, which maximizes the log-likelihood of the current token given a history of the ground truth. The dependence on ground truth leads to exposure bias \cite{ranzato2015sequence}, where the model can see the ground truth history during training. However, the ground truth is unavailable in testing so that the current prediction has to rely on its past predicted tokens. As a result, the incorrect predictions in the early time steps might result in error accumulation \cite{rennie2017self}.\par
There have been some previous works to alleviate the mismatches in the context of sequence-to-sequence models. The sampling-based method according to minimum Bayes risk (MBR) has been successfully applied to CTC model \cite{shannon2017optimizing} and RNN-T model \cite{weng2019minimum}. Furthermore, minimum WER (MWER) training has been proposed in attention-based model \cite{prabhavalkar2018minimum} and RNN-T model \cite{guo2020efficient}, which build MWER loss by sampling methods. However, these works only focus on the word error number of the entire utterance, while ignore the distinction of each token in a sequence.\par
Another alternative approach to address the mismatches is based on Reinforcement Learning (RL) since the sequence generation in ASR can be viewed as a sequential decision process in RL \cite{tjandra2018sequence}. The main idea of the RL-based method is to maximize the accumulative reward along all the time steps, and the customized reward function can build a direct link between the training objective and testing metric. Different from MWER training, a particular definition of reward function in RL is able to consider the impact of each token prediction on the entire generated sequence. In other words, MWER training is a case of RL that only considers the final reward of whole sequential decisions. In addition, RL-based methods have demonstrated their effectiveness in other sequence generation tasks, such like machine translation \cite{williams1992simple,bahdanau2016actor} and image caption \cite{rennie2017self}. \par
In this paper, we present a RL-based optimization method for sequence-to-sequence ASR task called self-critical sequence training (SCST). SCST associates the training loss and WER using WER-related reward function, which considers the intermediate reward at each token generation step. Furthermore, SCST utilizes the test-time beam search algorithm to sample a set of hypotheses for reward normalization. As a result, the high-reward hypotheses that outperform the current test-time system are given positive weights, while the low-reward hypotheses are given negative weights. In conclusion, the proposed SCST optimization method pushes the training procedure much closer to the test phase. The experiments on clean and noisy datasets show that it brings a relative improvement of 8.7\% and 7.8\% in terms of WER, respectively.
\vspace{-0.03in}
\section{Sequence-to-sequence ASR system}

In this section, we briefly introduce the general sequence-to-sequence ASR system from a decision-making perspective, and then explain the mismatches in this system. \par
Given a sequence of acoustic features $X = (x_1,x_2,...,x_N)$, the neural network in ASR system is expected to predict a sequence of tokens $Y = (y_1, y_2, ..., y_T)$, which is corresponding to the input acoustic sequence. Instead of directly outputting the tokens, the network usually predicts a possibility distribution of each token: $P_\theta(y_t | y_{t-1}, y_{t-2}, ... , y_0, X)$, where $t\in [0,T]$ refers to the $t$-th prediction. Accordingly, despite different structures of the network, two part of information are considered in $t$-th prediction: 1) The sequence of acoustic features $X$ and 2) previous generated tokens $Y_{t-1} = (y_{t-1},y_{t-2}, ... , y_0)$. Therefore, the sequence generation in ASR can be viewed as a sequential decision process along with $T$ time steps. Given the ground truth sequence $Y^* = (y^*_1, y^*_2, ..., y^*_T)$, most sequence-to-sequence models apply the cross-entropy criteria to calculate loss function:
\vspace{-0.05in}
\begin{equation}
\mathcal{L}_{ce} = \sum_{t=1}^T - \log  P_\theta(y_t^* |\ y_{t-1}^*, y_{t-2}^*, ... ,y_0^*,X)
\label{CEloss}
\end{equation}
\vspace{-0.1in}

The CE loss in Eq. (\ref{CEloss}) aims to maximize the log-likelihood for each token, while we typically use WER to evaluate the performance of trained model. However, WER is a non-differentiable task metric that can not be directly utilized in training loss. Therefore, the current training strategy suffers from this mismatch between training and testing. Furthermore, according to teacher-forcing algorithm, the prediction of current token relies on ground truth $Y*$. When ground truth is unavailable during testing, the incorrect predictions in an earlier time steps will accumulate errors through subsequent time steps.
\vspace{-0.03in}

\section{Reinforcement learning in ASR task}
In this section, we first illustrate how to model ASR task as a reinforcement learning problem. Secondly, the self-critical sequence training (SCST) method is proposed for ASR optimization. Finally, we discuss reward shaping with regards to sequence-to-sequence ASR.

\subsection{Connection between RL and ASR}
We first build the connection between the reinforcement learning formulation and ASR system. Basic reinforcement is modeled as a Markov decision process (MDP) which contains a tuple of ($S$, $A$, $P$, $R$) in successive time steps $t\in[0,T]$. The environment offers a current state $s_t\in S$. The agent takes $s_t$ into account and generates an instant action $a_t\in A$ which interacts with the environment. The $P_t$ denotes the transition probability from $s_t$ to $s_{t+1}$, and the $r_t$ refers the reward which is the feedback signal from environment. In reinforcement learning, the objective of training is to maximize the expected cumulative reward $R$: 
\vspace{-0.05in}
\begin{equation}
\mathbb{E}_{a_t\sim A} \ R =  \mathop{\max}_{a_t\in A} \sum_{t=0}^T  r_t
\label{reward}
\end{equation}

\begin{figure}[h!]
\centering
\includegraphics[width=0.49\textwidth]{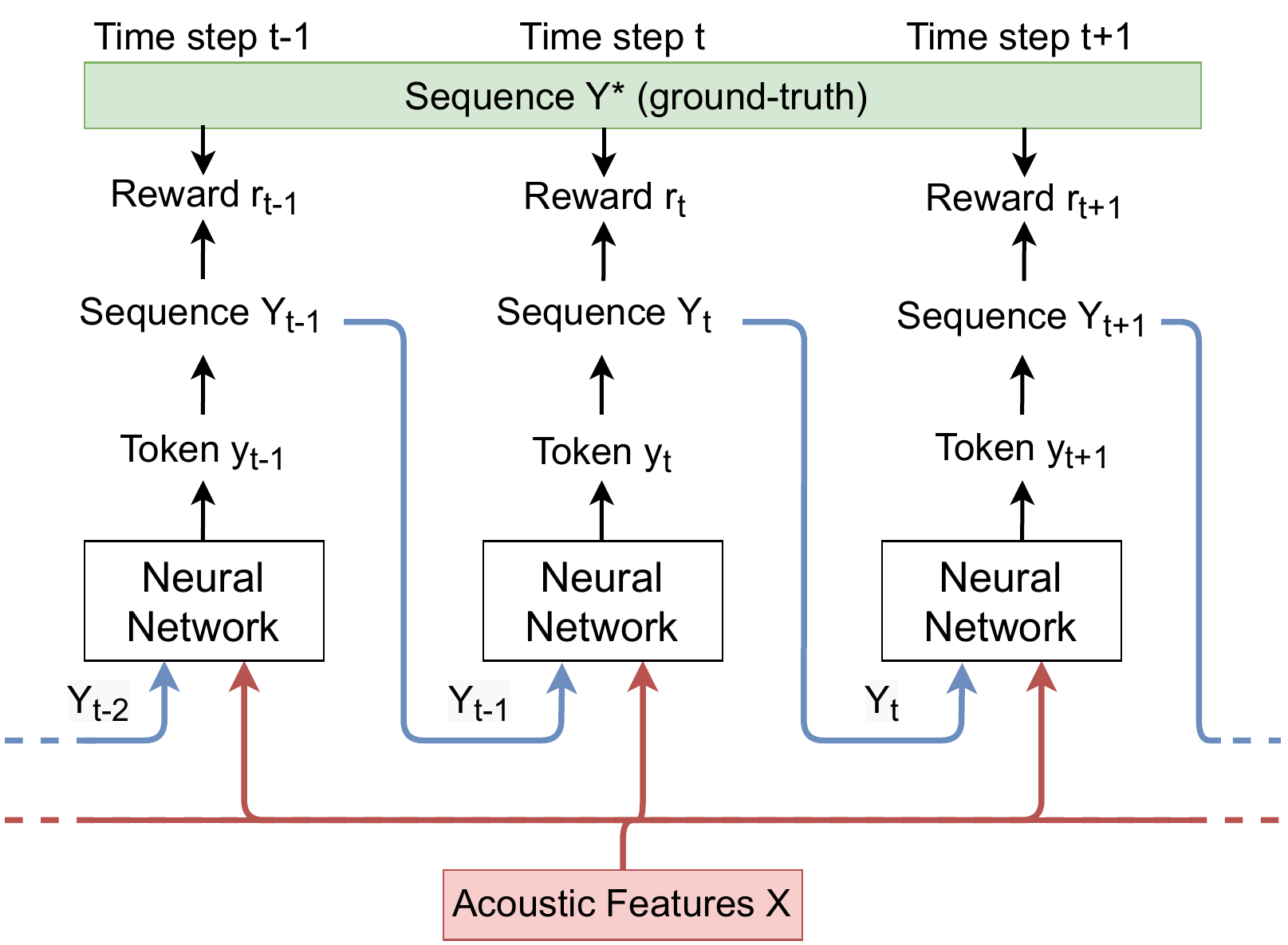}
\vspace{-0.25in}
\caption{The sequential decision model of ASR. Three temporal time steps are presented from right to left. For each time step, the neural network takes previous prediction $Y_{t-1}$ and acoustic features $X$ as input, generates a current token $y_t$, then updates the hypotheses sequence and calculated the reward $r_t$ by comparing with ground truth $Y^*$.}
\vspace{-0.1in}
\label{f1}
\end{figure}
For the sequence-to-sequence ASR task, it can also be viewed as a sequential decision model as shown in Fig. \ref{f1}. The whole encoder-decoder neural network can be viewed as an agent. In each time step $t$, acoustic feature $x_t$ and previous prediction $Y_{t-1}$ are concatenated as current state $s_t$. The output token is the action $a_t$ that will update the generated hypotheses sequence. After comparing it with ground truth sequence $Y^*$, a reward $r_t$ of this time step is calculated. Therefore, we define the training loss function to be the negative cumulative reward:
\vspace{-0.05in}
\begin{equation}
\mathcal{L}_\theta(X,Y^*) = - \mathbb{E}[R(Y,Y^*)] = \sum_Y P(Y|X,\theta) R(Y,Y^*) 
\label{rlloss01}
\end{equation}

where $R(Y,Y^*)=\sum_{t=0}^T r_t (Y_t,Y^*)$ refers to the reward of a hypotheses sequence that considers each time step from 0 to $T$. This is the difference between reinforcement learning and MWER training, where the latter only calculates the word error number of the entire hypotheses $Y$.

\subsection{Self-Critical Sequence Training}
In this part, we introduce a self-critical sequence training (SCST) approach and explain how it optimizes the ASR system using the N-best list. \par
In order to calculate the gradient $\nabla_\theta \mathcal{L}_\theta$ of Eq. (\ref{rlloss01}), the REINFORCE algorithms in \cite{williams1992simple} is employed to compute expected gradient of a non-differentiable reward function as follows:
\vspace{-0.05in}
\begin{equation}
\nabla_\theta \mathcal{L}_\theta = - \mathbb{E}_{Y^n\sim P(Y^n|X,\theta)}[R(Y^n,Y^*)\nabla_\theta log P(Y^n|X,\theta)] 
\label{rlloss02}
\end{equation}
where $Y^n$ = ($y_0^n$, $y_1^n$, ... , $y_T^n$) is hypotheses of sequence which is sampled from the current model. Instead of using other sampling method, we directly use N-best list \cite{bahdanau2016actor} hypotheses computed by Beam search decoding \cite{sutskever2014sequence} for the input utterance. Therefore, the number of samples is equal to beam size $N$, and we denote the $n$-th hypotheses as $ Y^n\in \text{Beam}(X,N) = (Y^1, Y^2, ... , Y^N)$. Furthermore, we introduce the \text{baseline} in \cite{sutton2018reinforcement} to normalize the rewards of the all hypotheses in N-best list:
\vspace{-0.05in}
\begin{equation}
\begin{small}
\nabla_\theta\mathcal{L}_\theta = -\frac{1}{N}\sum_{Y^n\in \text{Beam}}^N \nabla_\theta log P (Y^n|X,\theta) \ [\ R(Y^n,Y^*)-\Bar{R} \ ] 
\label{final_loss01}
\end{small}
\end{equation}
where $\Bar{R}$ is the \text{baseline}, and we define it as the average reward of hypotheses in $\text{Beam}(X,N)$. Subtracting $\Bar{R}$ from $R(Y^n,Y^*)$ does not change the expected gradient, but importantly, it can reduce the variance of the gradient estimate \cite{rennie2017self}. In order to simplify the calculation, we assume that the probability mass is concentrated on the N-best hypotheses only, thus the SCST loss $\mathcal{L}_{scst}$ could be approximated as:
\vspace{-0.05in}
\begin{equation}
\begin{small}
\mathcal{L}_{scst} \approx -\sum_{Y^n\in \text{Beam}}^N log\hat{P} (Y^n|X,\theta) \ [\ R(Y^n,Y^*)-\Bar{R} \ ] 
\label{final_loss02}
\end{small}
\end{equation}
where $\hat{P}(Y^n|X,\theta) = \frac{P(Y^n|X,\theta)}{\sum_{Y^n\in \text{Beam}} P(Y^n|X,\theta)}$ represents the re-normalized distribution over the N-best hypotheses, and $R(Y^n,Y^*) = \sum_{t=0}^T r_t$ has a temporal structure along the sequence. Eq. (\ref{final_loss02}) shows the central idea of the proposed SCST, which is to baseline the REINFORCE algorithm with the reward obtained by the current model using its inference mode. Accordingly, in an N-best list, the probability of hypotheses with higher reward than the average will be boosted, while the hypotheses which achieves lower reward will be suppressed. Therefore, in order to pursue higher reward, the SCST loss forces the trained model to explore the better WER performance using its inference mode.\par
In practice, the initial parameters of model is trained using CE loss $\mathcal{L}_{ce}$ in E.q (\ref{CEloss}). When $\mathcal{L}_{scst}$ is employed, we retrain the $\mathcal{L}_{ce}$ with a smaller weight $\lambda$. This operation is helpful to stabilize training for the sudden detachment of teacher-forcing. 

\subsection{Reward Shaping}
Reward shaping plays a significant role in almost all RL-related tasks. In this part, we discuss the set of reward functions $R(Y^n,Y^*)$ in the sequence-to-sequence ASR task. Since $R(Y^n,Y^*)$ is the medium building connection between training loss and testing metric, it is typically defined based on edit-distance which is directly related to WER.\par
\textbf{Reward  \uppercase\expandafter{\romannumeral1}}. Intuitively, the simplest reward function is to set rewards as negative edit-distance between a hypothesis and the ground truth, which is equal to MWER training. We denote the edit-distance between sequence $a$ and sequence $b$ as $ED(a,b)$, and the reward function for the hypotheses $Y^n$ the is shown as follows:
\vspace{-0.03in}
\begin{equation}
R_{\uppercase\expandafter{\romannumeral1}} (Y^n, Y^*) = -ED \ (Y^n,Y^*)
\label{reward1}
\end{equation}
where $Y^n$ denotes the $n$-th hypotheses, and $Y^*$ denotes the ground truth sequence.
\par
\textbf{Reward  \uppercase\expandafter{\romannumeral2}}.
Since reward  \uppercase\expandafter{\romannumeral1} only focus on the entire utterance of hypotheses, it ignores that the reward has a temporal structure along the sequence of predicted tokens. Therefore, we define the intermediate reward for each hypotheses $Y^n$ at each time step $t$ as $r_t(Y_t^n,Y_{t-1}^n,Y^*)$. Furthermore, we also retain the possibility history of each token $P_t(y_t|X, \theta)$, then multiply it by the reward for each token to calculate the temporal reward: 
\vspace{-0.07in}
\begin{equation}
\begin{aligned}
&R_{\uppercase\expandafter{\romannumeral2}} (Y^n, Y^*)= \sum_{t=0}^T \  r_t(Y_t^n,Y_{t-1}^n,Y^*)*P_t(y_t|X, \theta)
\\ 
&r_t(Y_t^n,Y_{t-1}^n,Y^*) = - [ED \ (Y_t^n,Y^*) - ED \ (Y_{t-1}^n,Y^*)]
\label{reward2}
\end{aligned}
\end{equation}
where $P_t(y_t|X, \theta)$ is the probability to predict the token $y_t$ at time step $t$. The $r_t(Y_t^n,Y_{t-1}^n,Y^*)$ is calculated to indicate whether the current new sequence $Y_t^n$ reduces the edit distance compared to previous sequence $Y_{t-1}^n$. \par
We notice that the reward  \uppercase\expandafter{\romannumeral2} may end up with a special case that the higher-reward hypotheses have more error words than lower-reward hypotheses. However, since the hypotheses in one N-best list are usually similar to each other, this case occurs with an extremely low probability, which means the higher reward can be approximated as lower WER. 

\section{Experiment Settings}
\subsection{Database}
We  conduct  experiments  on  the  dataset  from  robust  automatic transcription of speech (RATS) program \cite{graff2014rats}, which is recorded with a push-to-talk transceiver by playing back the clean Fisher data. The RATS has eight channels and could provide clean speech, noisy speech, and corresponding transcripts for various training goals. In this work, we choose the clean channel and channel A as clean and noisy conditions. They both include 44.3-hours of training data, 4.9-hours of validation data, and 8.2-hours of testing data \cite{chen2022noise}.

\subsection{Conformer-Based ASR Model}
we employed a Conformer-based ASR model \cite{gulati2020conformer} as the basic sequence-to-sequence system. The Conformer is a convolution-augmented Transformer \cite{vaswani2017attention}, which has achieved state-of-the-art performances on several public datasets. Furthermore, in order to obtain better performance, the end-to-end ASR model is jointly trained using both CTC \cite{ma2021multitask} and attention-based cross-entropy criteria.\par
In practice, the Conformer-based ASR system takes the 80-dim Log-Mel feature as input. The encoder contains 12 Conformer layers, while the decoder consists of 6 Transformer layers. We use 994 byte-pair-encoding (BPE) \cite{kudo2018sentencepiece} tokens as ASR output. We set the weight of CTC loss to 0.3, and train 55 epochs using Adam optimizer with an initial learning rate of 0.002. The best-performance model is selected as a baseline using the validation set. We then use this model for N-best sampling and start to apply the SCST optimization. For a fair comparison, the SCST training will also end at epoch 55.
\vspace{-0.05in}
\section{Results}
\subsection{Effect of the CE Loss Weight}
We first analyse the importance of CE loss weight $\lambda$ when using SCST optimization on the clean RATS dataset. We fix the N-best list size at 5 and apply reward  \uppercase\expandafter{\romannumeral1}, and gradually increase the value of $\lambda$ from 0. From Table \ref{table1}, we observe that it is important to interpolate with CE loss during optimization. When $\lambda$ = 0, the ASR model performs even worse due to a sudden change of training loss. 
However, with a small weight of $\lambda$, the SCST optimization surpasses the baseline model that is only trained with CE loss, and the best performance is obtained with $\lambda=0.001$. \par

\begin{table}[t]
\centering
\caption{The comparative study of different CE loss weight $\lambda$ during optimization on clean RATS dataset. }
\begin{tabular}{c|c|c}
\hline\hline
model    & $\lambda$       & WER (\%)\\ \hline\hline
Baseline & -       &   29.9                                                 \\ \hline
SCST    & 0       &   34.8                                                 \\ \hline
SCST    & 0.0001  &   28.1                                                 \\  \hline

SCST    & 0.001   &   \textbf{28.0}                                               \\ \hline
SCST    & 0.01    &   28.3                                                \\ 
\hline\hline
\end{tabular}
\vspace{-0.25in}
\label{table1}
\end{table}

\subsection{Effect of Two Type of Rewards}
We then explore the result of different reward functions. Two reward functions are employed in SCST optimization onto the same baseline. Since CE loss weight has a slight influence on results, two values of $\lambda$ are utilized in each reward type to explore better WER. Furthermore, the character error rate (CER) are also reported, because reward \uppercase\expandafter{\romannumeral2} is a token-level reward function.\par  
\begin{table}[t]
\centering
\caption{The comparative study of two reward types during SCST optimization on clean RATS dataset.}
\begin{tabular}{c|c|c|c|c}
\hline\hline
Model    & $\lambda$ & Reward                      & WER (\%)              & CER (\%) \\ \hline
Baseline & -      & -                     & 29.9                  & 19.5     \\ \hline\hline
SCST     & 0.001 & Reward \uppercase\expandafter{\romannumeral1}   &   28.0 & 18.2   \\ \hline
SCST     & 0.001 & Reward \uppercase\expandafter{\romannumeral2}     &  27.5  & 17.7    \\ \hline
SCST     & 0.0001 & Reward \uppercase\expandafter{\romannumeral1}   &  28.1  & 18.4     \\ \hline
SCST     & 0.0001  & Reward \uppercase\expandafter{\romannumeral2}    &  \textbf{27.3} & \textbf{17.6}  \\ \hline\hline
\end{tabular}
\vspace{-0.15in}
\label{table2}
\end{table}
From Table \ref{table2}, we observe that optimization with two types of rewards both surpass the baseline. The \uppercase\expandafter{\romannumeral2} achieves the better performance, which obtains 8.7\% WER improvement and 9.7\% CER improvement, both relatively.     
\subsection{Generalization on Noisy Dataset}
we also report the SCST optimization on the RATS channel A dataset, which contains very noisy data over the radio. Since the noisy speech is more difficult to recognize, the value of expected reward reduces when employing SCST optimization. Therefore, we reduce the CE weight $\lambda$, and the result is shown in Table \ref{table3}. Similar to results on the clean RATS dataset, all optimized models by SCST surpass the baseline model in terms of WER and CER. The best model obtains 7.8\% relative WER improvement and 8.6\% relative CER improvement, respectively. However, with the increase of the recognition difficulty, Reward \uppercase\expandafter{\romannumeral2} shows no more superiority over Reward \uppercase\expandafter{\romannumeral1} on the noisy dataset.
\begin{table}[t]
\centering
\caption{The comparative study of two reward types during SCST optimization on noisy RATS channel A dataset.}
\begin{tabular}{c|c|c|c|c}
\hline\hline
Model    & $\lambda$ &  Reward                     & WER (\%)              & CER (\%) \\ \hline
Baseline & -      & -                     & 58.7                  & 42.9     \\ \hline\hline
SCST     & 0.0001 & Reward \uppercase\expandafter{\romannumeral1}   &   54.1 & 39.2   \\ \hline
SCST     & 0.0001 & Reward \uppercase\expandafter{\romannumeral2}     &  54.3  & 39.4  \\ \hline
SCST     & 0.00001 & Reward \uppercase\expandafter{\romannumeral1}   &  54.2  & 39.5   \\ \hline
SCST    & 0.00001 & Reward \uppercase\expandafter{\romannumeral2}     &  54.1 & 39.2  \\ \hline\hline
\end{tabular}
\label{table3}
\vspace{-0.15in}
\end{table}
\section{Conclusion}
In this paper, we proposed a SCST optimization method to address the mismatch problems between training and testing in ASR task. This RL-based method builds a direct link between training objective and testing metric and also harmonizes the model with respect to its inference procedure. The results show that SCST optimization is effective on both clean dataset and noisy dataset, which respectively obtains 8.7\% and 7.8\% relative WER improvements over baseline that is only trained using CE loss.

\vfill\pagebreak

\bibliographystyle{IEEEtran}
\bibliography{refs}

\end{document}